\documentclass[letterpaper, 10 pt, conference]{ieeeconf}  

\IEEEoverridecommandlockouts                              

\title{\LARGE \bf
Situated Haptic Interaction: Exploring the Role of Context in Affective Perception of Robotic Touch}

\author{Qiaoqiao Ren and Tony Belpaeme
\thanks{Qiaoqiao Ren and Tony Belpaeme are with AIRO-IDLab, Department of Electronics and Information Systems, Ghent University - imec
        {\tt\small Qiaoqiao.Ren@ugent.be}
}
}

\usepackage{graphicx}
\usepackage{subcaption}
\usepackage{color}
\usepackage{textcomp}
\usepackage{csquotes}
\usepackage{hyperref}

\usepackage{caption} 
\usepackage{tabularx}
\usepackage{float}
\usepackage{array}  
\usepackage{booktabs} 
\usepackage{multirow}
\usepackage{makecell}
\usepackage{arydshln} 
\usepackage{todonotes}
\usepackage{eurosym}
\usepackage{balance}

\usepackage{siunitx}
\usepackage{cite}
\begin{document}


\maketitle

\begin{abstract}

Affective interaction is not merely about recognizing emotions; it is an embodied, situated process shaped by context and co-created through interaction. In affective computing, the role of haptic feedback within dynamic emotional exchanges remains underexplored. This study investigates how situational emotional cues influence the perception and interpretation of haptic signals given by a robot.
In a controlled experiment, 32 participants watched video scenarios in which a robot experienced either positive actions (such as being kissed), negative actions (such as being slapped) or neutral actions. After each video, the robot conveyed its emotional response through haptic communication, delivered via a wearable vibration sleeve worn by the participant. Participants rated the robot’s emotional state—its valence (positive or negative) and arousal (intensity)—based on the video, the haptic feedback, and the combination of the two.
The study reveals a dynamic interplay between visual context and touch. Participants’ interpretation of haptic feedback was strongly shaped by the emotional context of the video, with visual context often overriding the perceived valence of the haptic signal. Negative haptic cues amplified the perceived \textit{valence} of the interaction, while positive cues softened it. Furthermore, haptics override the participants’ perception of \textit{arousal} of the video.
Together, these results offer insights into how situated haptic feedback can enrich affective human-robot interaction, pointing toward more nuanced and embodied approaches to emotional communication with machines.
\end{abstract}

\section{Introduction}

Emotions are central to human experience, shaping how we engage with others and interpret the world around us. In the fields of Human-Computer Interaction (HCI) and Human-Robot Interaction (HRI), there is increasing recognition that emotions are not merely internal cognitive states but are context-dependent, embodied, and dynamically constructed through interactions \cite{dourish2001action, battarbee2004co}. Traditional affective computing has largely focused on recognizing and categorizing emotions based on predefined models, such as Ekman’s six basic emotions \cite{ekman1999basic} or Russell’s circumplex model \cite{russell1980circumplex}. However, these models often oversimplify emotional experiences, failing to capture the nuanced and emergent nature of emotions in real-world interactions. Recent research in affective interaction challenges these reductionist frameworks, emphasizing that emotions emerge through bodily experiences, social context, and multimodal interactions \cite{ahmadpour2025affective}.

One underexplored modality in affective interaction is touch \cite{eid2015affective}. While vision and auditory cues have been dominant in affective computing research, haptic feedback plays a crucial role in emotional communication \cite{van2015social}. Humans frequently use touch to express and perceive emotions, from a comforting pat on the back to a gentle stroke or a reassuring squeeze \cite{suvilehto2023and}. Psychological and neuroscientific research suggests that affective touch is fundamental to social bonding, emotional regulation, and communication \cite{gallace2014touch}. Studies in human-human interaction have demonstrated that different types of touch (e.g., stroking, pressing, squeezing, tapping) convey distinct emotional meanings \cite{app2011nonverbal}. 

In Human-Robot Interaction (HRI), affective touch has been explored primarily in therapeutic and assistive applications \cite{yohanan2012role}. For example, socially assistive robots such as PARO (a robotic seal designed for elderly care) provide comfort through tactile interaction \cite{wada2007living}. However, much of the existing research on haptic feedback in robotics has focused on sensory augmentation rather than emotional communication \cite{shull2015haptic}. Some studies have investigated how robots can convey emotions through haptic modalities, varying touch intensity, duration, and rhythm \cite{ yohanan2012role}. And previous research indicates that human-robot tactile interaction could influence people's behaviour and attitudes \cite{ren2024tactile, ren2023behavioural}. Furthermore, emotional cues in visual and haptic modalities have been examined in relation to multimodal emotion perception \cite{akshita2015towards}. These studies suggest that haptic signals can effectively communicate emotions, but they do not fully address the situational and contextual factors that influence the interpretation of affective touch.

With advancements in haptic technology, researchers have begun exploring mediated touch—where tactile signals are transmitted through digital or robotic interfaces—as a means of conveying emotions. Mediated touch systems enable remote affective communication, such as haptic telepresence, where one user’s touch is reproduced through a robotic or vibrotactile interface \cite{smith2007communicating}. Vibrotactile haptics can encode emotions through variations in frequency, amplitude, and rhythm, simulating natural touch-like sensations \cite{rantala2013touch}. Previous research has suggested that certain vibrotactile patterns can be associated with positive and negative emotions \cite{ju2021haptic, ren2025touched}. This is particularly relevant in robotic interaction, where direct human-like touch is not always possible, and vibrations can serve as an alternative means of emotional expression.

Affective perception is not solely determined by the physical properties of haptic feedback; it is also shaped by situational cues and contextual meaning \cite{hoemann2017mixed}. In human-human interaction, the same physical touch can carry vastly different meanings depending on the context \cite{gallace2010science}. Moreover, more recent studies indicate that contextual and social factors significantly influence the emotional interpretation of touch \cite{van2015social}. For instance, a firm handshake in a business setting conveys professionalism and confidence, whereas the same touch in an intimate setting may feel cold and distant \cite{huwer2003understanding}. Similarly, research has shown that most touch behaviours carry symbolic meaning and can effectively communicate complex emotions \cite{jones1985naturalistic}. In HRI, similar principles apply—how people interpret a robot’s haptic feedback depends on the situational cues surrounding the interaction \cite{urakami2023nonverbal}. Multimodal signals—such as facial expressions, verbal utterances, and body language—significantly shape how people perceive touch-based robotic interactions \cite{tielman2014adaptive}. However, existing studies have not systematically examined the role of context in shaping the perception of robotic haptic feedback. 

To address this gap, this study aimed to investigate how situational context and tactile feedback interact to influence emotional interpretation during interactions with a humanoid robot (Pepper). Specifically, we focus on the following research questions (RQs):

\begin{enumerate}
    \item \textbf{RQ1}: Can participants decode emotions through vibration by a mediated wearable haptic device?
    \item \textbf{RQ2}: Does situational context override or bias the interpretation of a haptic stimulus's valence? 
    \item \textbf{RQ3}: Does situational context override or bias the interpretation of a haptic stimulus's arousal? 
\end{enumerate}

\section{Methods}


\subsection{Participants}

Thirty-two Chinese participants (10 female, 18 male; M = 27.8, SD = 2.3 years) participated in the experiment. Participants were recruited from a similar cultural background to ensure consistency in emotional interpretation. The study complied with ethical guidelines established by \emph{Ghent University}, and informed consent was obtained from all.

\subsection{Experimental design}

The experiment employed a within-subject design, where each participant experienced all conditions, including: (1) two types of haptic feedback (comfort stimulus with positive valence and low arousal, and anger stimulus with negative valence and high arousal \cite{russell1980circumplex}); (2) six situational context videos (Kiss, Slap, Eye Contact, Stroke, Flick, Cover Eyes), which is available in our GitHub repository\footnote{\url{https://github.com/qiaoqiao2323/Situational_haptic_interaction/tree/main}}; and (3) the combination of each video with both types of haptic feedback. Participants were asked to decode the emotions conveyed by the Pepper robot (present in the experiment room) through a vibration sleeve \cite{ren2025touched}, as shown in Fig.~\ref{fig:motors}.

\subsection{Haptic data construction}

We conducted data collection experiments to collect touch data associated with emotions through a $5 \times 5$ piezoresistive tactile sensor, as described in \cite{ren2024conveying}. Participants first expressed ten different emotions by touching the Pepper robot's upper arm, resulting in a dataset of 84 tactile files in CSV format at 45 Hz, each 10 seconds in duration.

To analyse this data, for each emotion, we applied the k-means clustering algorithm, grouping the audio features and tactile features in \cite{ren2024conveying} into three clusters based on their acoustic features. We identified the most populated cluster, representing the largest grouping of participants, as the most characteristic. Within this cluster, we calculated the Euclidean distance of each participant's expression features to the cluster centroid (average feature values) to assess each clip's representativeness. The three participants' IDs with the smallest distances to the centroid were selected as the most representative, and we chose one participant's data among the three to serve as a representative stimulus for all participants. In this experiment, we selected two haptic stimuli based on participant expressions: one representing ``anger'', characterized by high arousal and negative valence, and the other representing ``comfort'', characterized by low arousal and positive valence.

Then we translate the tactile files to vibrations and the system uses pulse-width modulation (PWM) for precise control over the intensity of each motor’s vibration, which allows for modulation of the vibration intensity by rapidly switching the transistors on and off at varying duty cycles. Each vibration stimulus lasts for 10 seconds as well.


\subsection{Situational context actions construction}

Inspired by \cite{read2014situational}, we designed six distinct situational context interactions: Kiss, where the Pepper robot received a kiss from the human; Slap, where the human slapped Pepper’s face; Eye Contact, where Pepper attempted to establish eye contact by looking at the camera; Stroke, where the human gently stroked Pepper’s head; Flick, where the human flicked Pepper’s forehead; and Cover Eyes, where the human covered Pepper’s eyes.

\subsection{Procedures}

\begin{figure}[h]
    \centering
        \includegraphics [height=3.9cm]{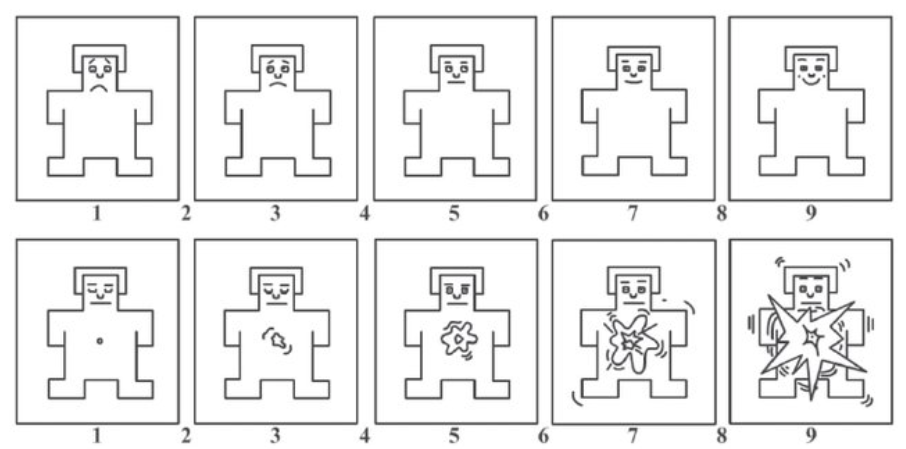}
    \caption{SAM for valence and arousal, the first row is valence and the second is arousal.}
    \label{fig::sam}
\end{figure}

\begin{figure*}
\centering
\begin{subfigure}{0.5\textwidth}
    \centering
    \includegraphics[height=5cm]{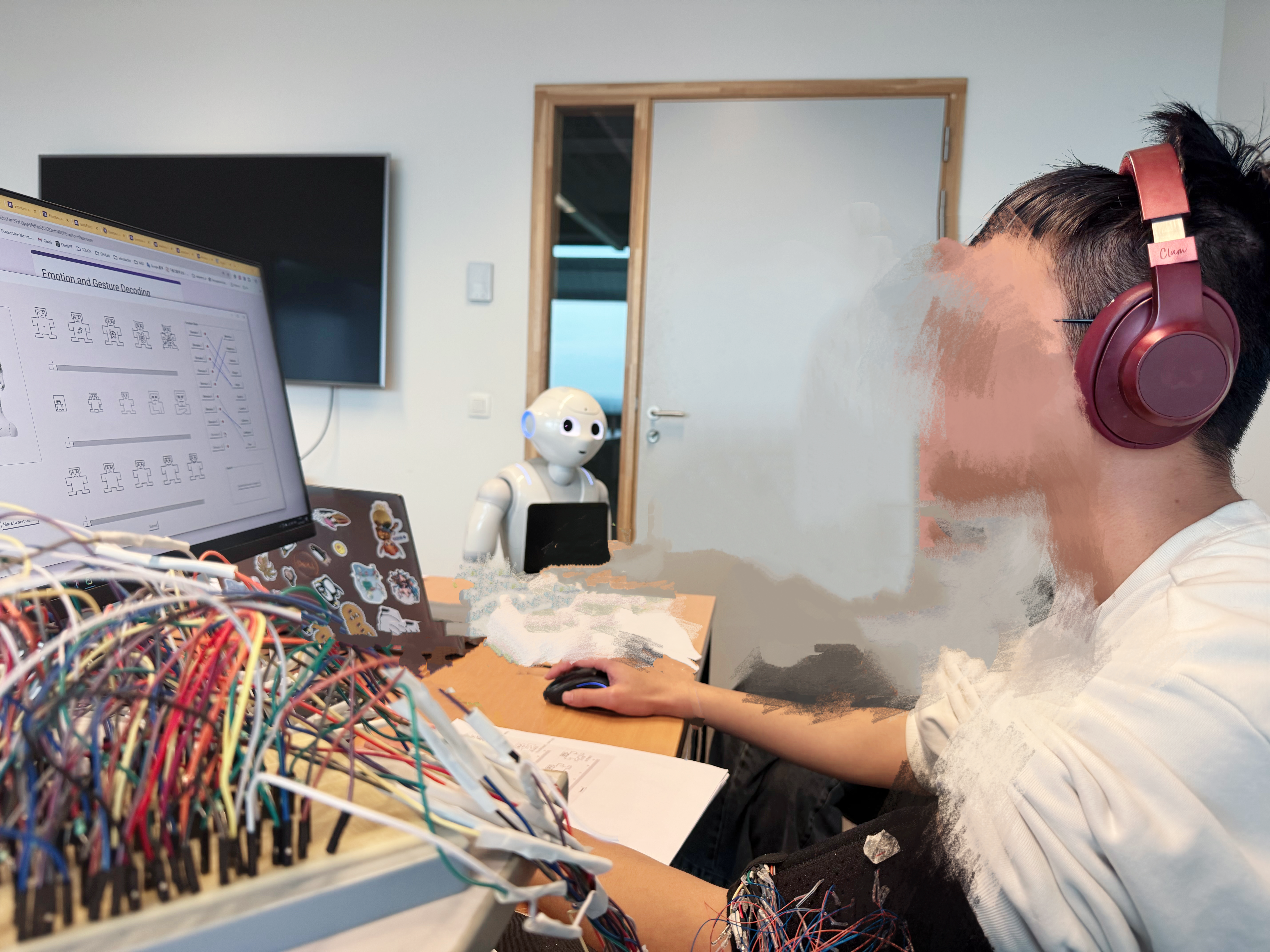}
\caption{Participants decoding emotions.}
    \label{fig:interaction}
\end{subfigure}
\begin{subfigure}{0.45\textwidth}
    \centering
    \includegraphics[height=5cm]{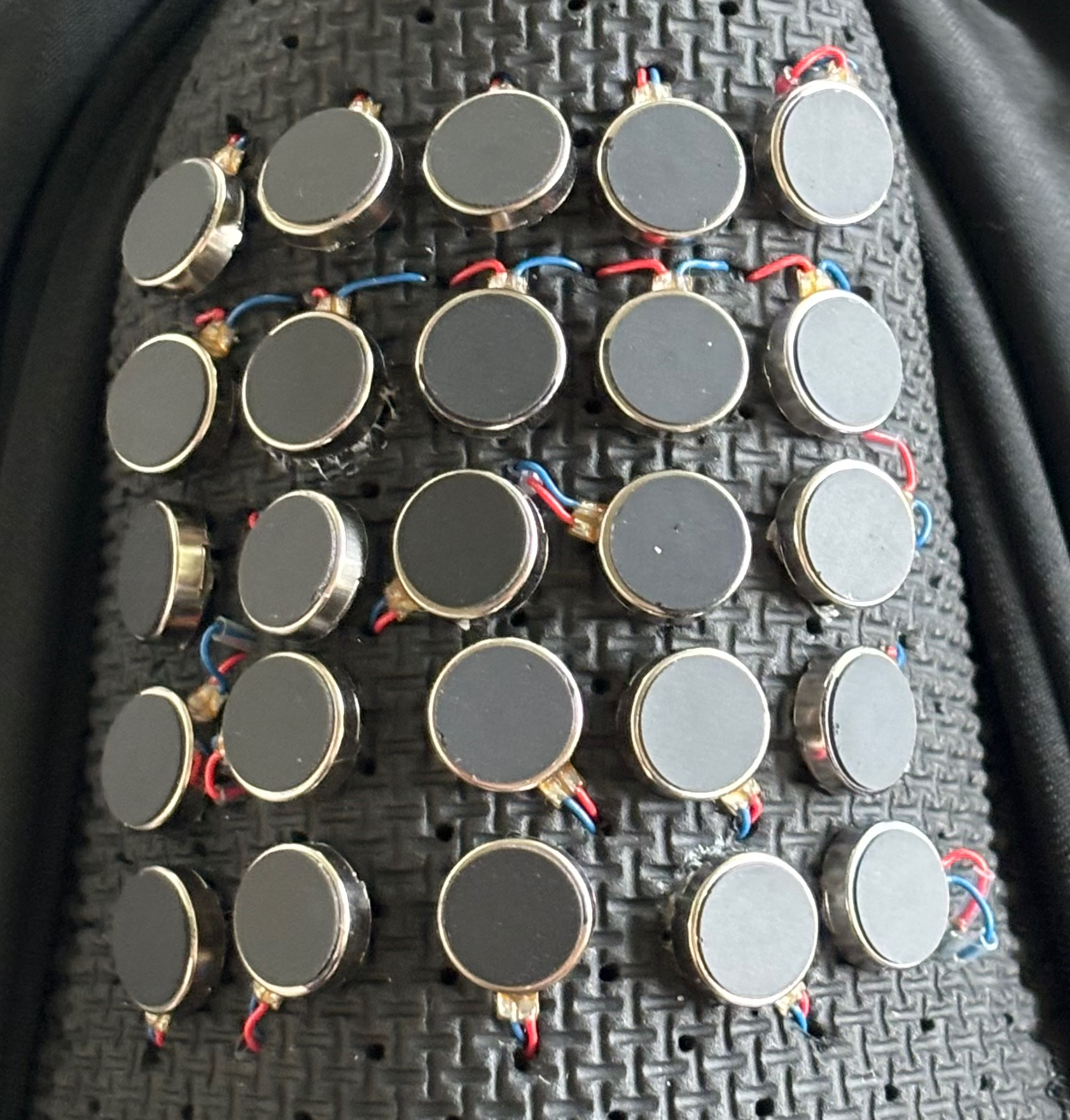}
    \caption{Distribution of vibration motors.}
    \label{fig:motors}
\end{subfigure}
\caption{Vibration sleeves.}
\label{fig:experimentalsetup}
\end{figure*}

Each participant underwent four phases as shown in Fig.~\ref{fig:interaction}; in each phase, they were asked to decode emotions for each stimulus by Self Assessment Manikins (SAM) as shown in Fig.~\ref{fig::sam}.

\vspace{-3mm}

\begin{enumerate}
    \item \textbf{Pre-session (Threshold Perception)}: It served to ensure that all participants could perceive the vibrations.

    \item \textbf{Tactile-only Evaluation}: In this phase, the robot delivers an anger stimulus (\textbf{C\textsubscript{Tactile\textsuperscript{Anger}}}) and comfort (\textbf{C\textsubscript{Tactile\textsuperscript{Comfort}}}) stimulus, and participants need to rate the arousal and valence of each stimulus.

    \item \textbf{Context-only Evaluation}: Participants rated emotional states based solely on observing video scenarios (C\textsubscript{Context}). In this session, participants observe the robot experiencing an action from a human (e.g., being kissed or slapped) without receiving tactile feedback. Specifically, including \textbf{C\textsubscript{Kiss}}, \textbf{C\textsubscript{Slap}}, \textbf{C\textsubscript{Eye Contact}}, \textbf{C\textsubscript{Stroke}}, \textbf{C\textsubscript{Flick}}, \textbf{C\textsubscript{Cover eyes}}.

    \item \textbf{Combined Tactile and Context Evaluation}: Participants rated arousal and valence while observing video scenarios followed by vibration feedback sending from the robot C\textsubscript{Context\textsuperscript{Anger}}, C\textsubscript{Context\textsuperscript{Comfort}}).

        \begin{itemize}
            
            \item Participants observe the robot experiencing an action and responding with an ``Anger'' vibration C\textsubscript{Context\textsuperscript{Anger}}, as well as a ``Comfort'' vibration C\textsubscript{Context\textsuperscript{Comfort}}, here the Context includes Slap, Eye Contact, Stroke, Flick, Cover eyes and Kiss. 
                    
        \end{itemize}

\end{enumerate}

\section{Results and analysis}



\begin{table*}[t]\footnotesize
\setlength{\abovecaptionskip}{0.0cm}   
\setlength{\belowcaptionskip}{-0cm}  
\renewcommand\tabcolsep{2.0pt} 
\centering
\caption{Arousal scores for situational context affective tactile interaction.}
\begin{tabular}{
p{3cm}<{\centering}  
p{2cm}<{\centering} p{2cm}<{\centering}  
p{2cm}<{\centering} p{2cm}<{\centering}  
p{2cm}<{\centering} p{2cm}<{\centering}  
}
\hline
{Haptic Stimulus} & Eye Contact & Cover Eyes & Flick & Kiss & Slap & Stroke \\
\hline
No haptic feedback & $3.25\pm1.81$ & $6.25\pm2.03$ & $5.75\pm1.98$ & $5.84\pm2.49$ & $7.81\pm1.45$ & $3.78\pm2.39$ \\
Anger ($8.06\pm1.50)$ & $7.09\pm1.99$ & $8.13\pm1.24$ & $7.81\pm1.33$ & $8.25\pm1.19$ & $8.28\pm1.28$ & $7.28\pm1.85$ \\
Comfort ($4.16\pm1.71$ & $4.16\pm1.71$  & $5.41\pm1.72$ & $4.84\pm1.82$& $5.19\pm2.10$ & $5.75\pm2.06$ & $4.93\pm1.83$ \\
\hline
\end{tabular}
\label{tab:arousal}
\end{table*}

\begin{table*}[t]\footnotesize
\setlength{\abovecaptionskip}{0.0cm}   
\setlength{\belowcaptionskip}{-0cm}  
\renewcommand\tabcolsep{2.0pt} 
\centering
\caption{Valence scores for situational context affective tactile interaction.}
\begin{tabular}{
p{3cm}<{\centering}  
p{2cm}<{\centering} p{2cm}<{\centering}  
p{2cm}<{\centering} p{2cm}<{\centering}  
p{2cm}<{\centering} p{2cm}<{\centering}  
}
\hline
{Haptic Stimulus} & Eye Contact & Cover Eyes & Flick & Kiss & Slap & Stroke \\
\hline
No haptic feedback & $5.63\pm1.29$ & $3.25\pm1.67$ & $3.28\pm1.40$ & $8.13\pm0.91$ & $1.72\pm1.05$ & $6.78\pm1.31$ \\
Anger ($2.06\pm1.41)$ & $5.28\pm2.30$ & $2.72\pm1.76$ & $2.97\pm1.64$ & $7.94\pm1.70$ & $1.50\pm0.72$ & $7.38\pm1.54$ \\
Comfort ($5.09\pm1.94$) & $5.16\pm1.35$ & $3.94\pm1.72$ & $3.69\pm1.31$ & $6.43\pm1.32$ & $2.38\pm1.01$ & $6.25\pm1.52$ \\

\hline
\end{tabular}
\label{tab:valence}
\end{table*}

\subsection{Tactile stimulus only}

A Wilcoxon rank-sum test indicated significant differences between C\textsubscript{Tactile\textsuperscript{Anger}} ($8.06\pm1.50$)  and C\textsubscript{Tactile\textsuperscript{Comfort}} ($4.16\pm1.71$) haptic stimuli for arousal ratings, ($W = 951, p < .001$). Similarly, valence ratings significantly differed between C\textsubscript{Tactile\textsuperscript{Anger}}  ($2.06\pm1.41$) and C\textsubscript{Tactile\textsuperscript{Comfort}} ($5.09\pm1.94$), ($W = 71, p < .001$). The results are shown in Table.~\ref{tab:arousal}. This confirmed that participants can decode emotions through vibration sent by Pepper robots.

\subsection{Action videos}
\subsubsection{Arousal analysis}

The results are shown in the Tab.~\ref{tab:arousal} and Tab.~\ref{tab:valence}. The Kruskal-Wallis test revealed a significant main effect of situational context on arousal ($H(5) = 69.68, p < 0.001$). Post-hoc Wilcoxon signed-rank tests with Bonferroni correction further explored these differences. For arousal ratings, significant differences emerged notably between the \textbf{C\textsubscript{Cover eyes}} and \textbf{C\textsubscript{Eye Contact}} conditions ($p<0.001$), as well as between the \textbf{C\textsubscript{Cover eyes}} and \textbf{C\textsubscript{Stroke}} conditions ($p=0.0015$), and  \textbf{C\textsubscript{Cover eyes}} versus  \textbf{C\textsubscript{Slap}} conditions ($p=0.0194$). Additionally, the  \textbf{C\textsubscript{Flick}} scenario significantly differed from  \textbf{C\textsubscript{Eye Contact}} ($p<0.001$),  \textbf{C\textsubscript{Slap}} ($p<0.001$), and  \textbf{C\textsubscript{Stroke}} conditions ($p=0.0111$). Significant differences were also observed between the  \textbf{C\textsubscript{Kiss}} scenario and  \textbf{C\textsubscript{Eye Contact}} ($p<0.001$),  \textbf{C\textsubscript{Slap}} ($p=0.0067$), and \textbf{C\textsubscript{Stroke}} ($p=0.0450$) conditions. Lastly, the  \textbf{C\textsubscript{Slap}} scenario was significantly different from the \textbf{C\textsubscript{Stroke}} ($p<0.001$). However, there is no significant difference between \textbf{C\textsubscript{Cover eyes}} and \textbf{C\textsubscript{Flick}}, as well as between \textbf{C\textsubscript{Cover eyes}} and \textbf{C\textsubscript{Kiss}}; in addition, no significant difference was found between \textbf{C\textsubscript{Flick}} and \textbf{C\textsubscript{Kiss}}, and \textbf{C\textsubscript{Eye Contact}} and \textbf{C\textsubscript{Stroke}} on arousal, which indicate that some situational context has similar arousal level. These results indicate robust variations in participants' arousal responses depending on the situational context.

\subsubsection{Valence analysis}

The Kruskal-Wallis test revealed a significant main effect of situational context on valence ratings ($ H(5) = 145.81, p < 0.001$). Post-hoc Wilcoxon signed-rank tests with Bonferroni correction further explored these differences. Significant differences emerged notably between the \textbf{C\textsubscript{Cover eyes}} and \textbf{C\textsubscript{Kiss}} conditions ($p < 0.001$), as well as between \textbf{C\textsubscript{Cover eyes}} and \textbf{C\textsubscript{Eye Contact}} ($p < 0.001$), and \textbf{C\textsubscript{Cover eyes}} versus \textbf{C\textsubscript{Stroke}} ($p < 0.001$). Additionally, the \textbf{C\textsubscript{Flick}} scenario significantly differed from \textbf{C\textsubscript{Kiss}} ($p < 0.001$),  \textbf{C\textsubscript{Eye Contact}} ($p < 0.001$),  \textbf{C\textsubscript{Slap}} ($p < 0.001$), and  \textbf{C\textsubscript{Stroke}} ($p < 0.001$) conditions. Significant differences were also observed between the  \textbf{C\textsubscript{Kiss}} scenario and \textbf{C\textsubscript{Eye Contact}} ($p < 0.001$), \textbf{C\textsubscript{Slap}} ($p < 0.001$), and \textbf{C\textsubscript{Stroke}} ($p = 0.0007$) conditions. Lastly, the \textbf{C\textsubscript{Slap}} scenario was significantly different from both \textbf{C\textsubscript{Eye Contact}} ($p < 0.001$) and \textbf{C\textsubscript{Stroke}} ($p < 0.001$). These results indicate a significant difference in participants' valence responses across all the situational contexts, highlighting that the proposed six situational contexts elicit significantly different valence.


\begin{figure*}
\centering
\begin{subfigure}{0.5\textwidth}
    \centering
    \includegraphics[width=\linewidth]{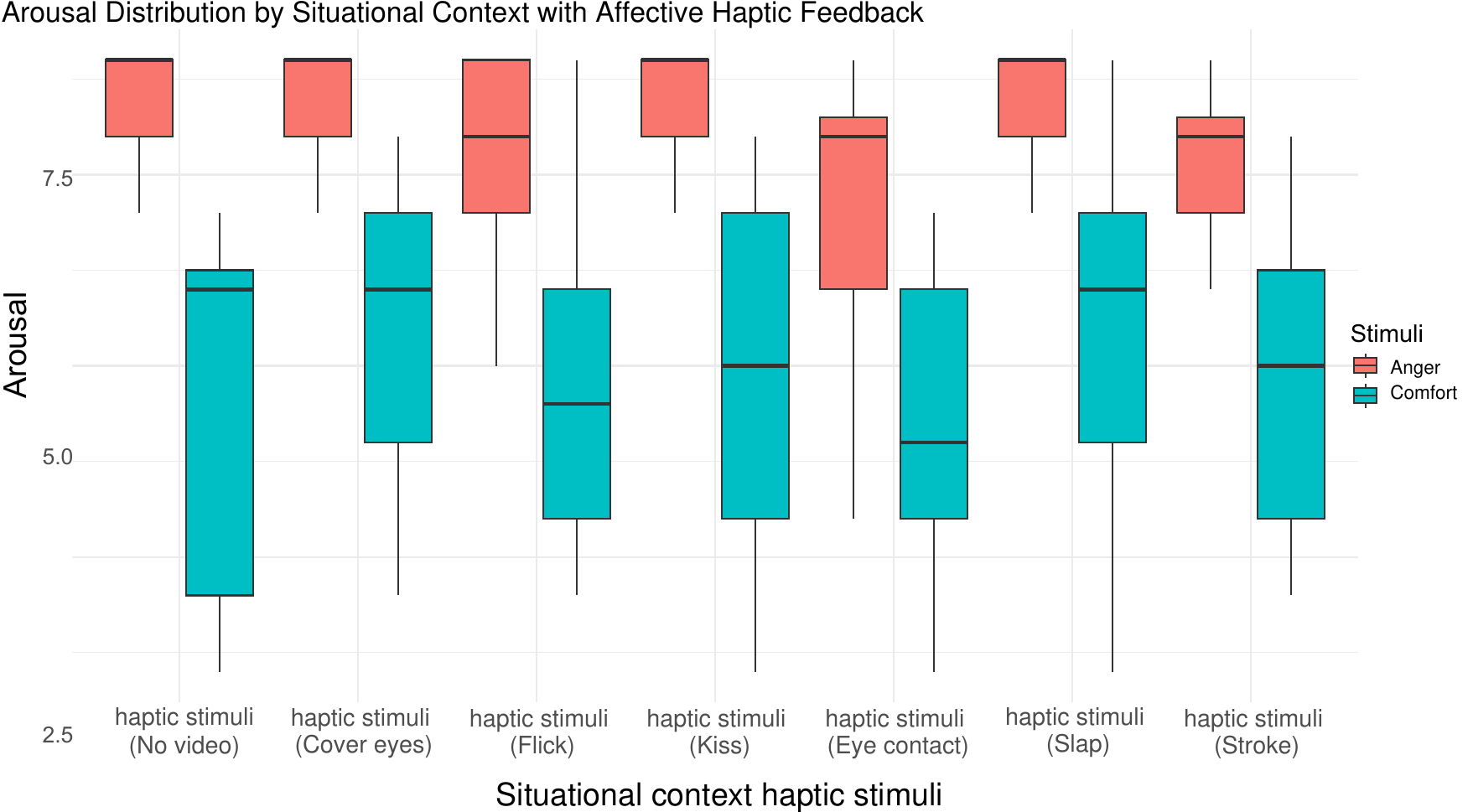}
\caption{Arousal comparison between the tactile feedback only and video with tactile feedback, blue bars represent ``Comfort'' stimulus and red bars are with ``Anger'' stimulus. }
    \label{fig:touch_arousal}
\end{subfigure}
\begin{subfigure}{0.45\textwidth}
    \centering
    \includegraphics[width=\linewidth]{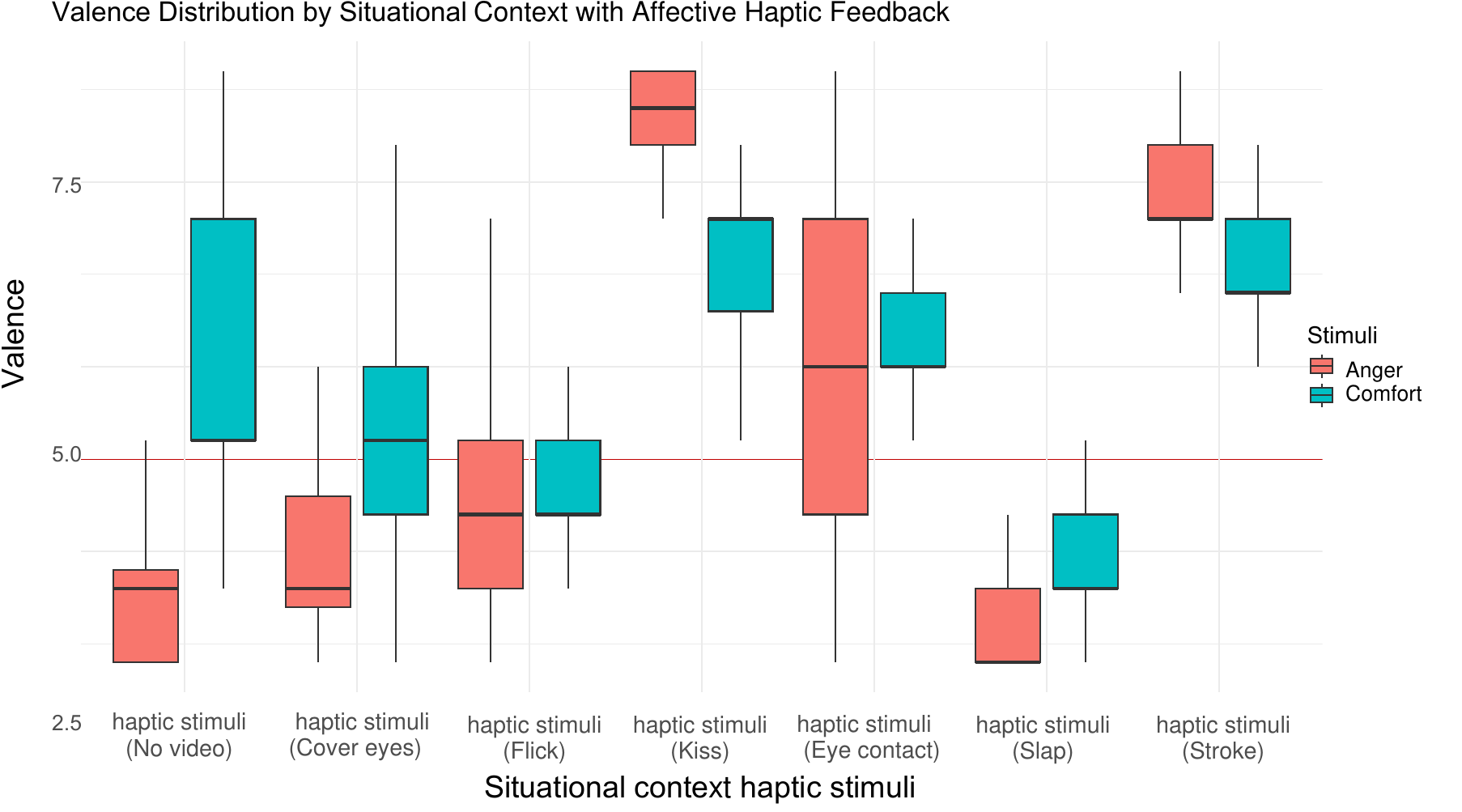}
    \caption{Valence comparison between the tactile feedback and video with tactile feedback, blue bars represent ``Comfort'' stimulus and red bars are with ``Anger'' stimulus. }
    \label{fig:touch_valence}
\end{subfigure}
\caption{Comparison between the tactile feedback only and video with tactile feedback. In subfigure (a), it's shown that the haptic stimulus arousal overrides the situational context arousal and subfigure (b) shows that negative valence haptic stimulus exaggerates the valence while the positive emotion modulates the valence, converging to the neutral emotion.}
\label{fig:touch}
\end{figure*}

\begin{figure*}
\centering
\begin{subfigure}{0.5\textwidth}
    \centering
    \includegraphics[width=\linewidth]{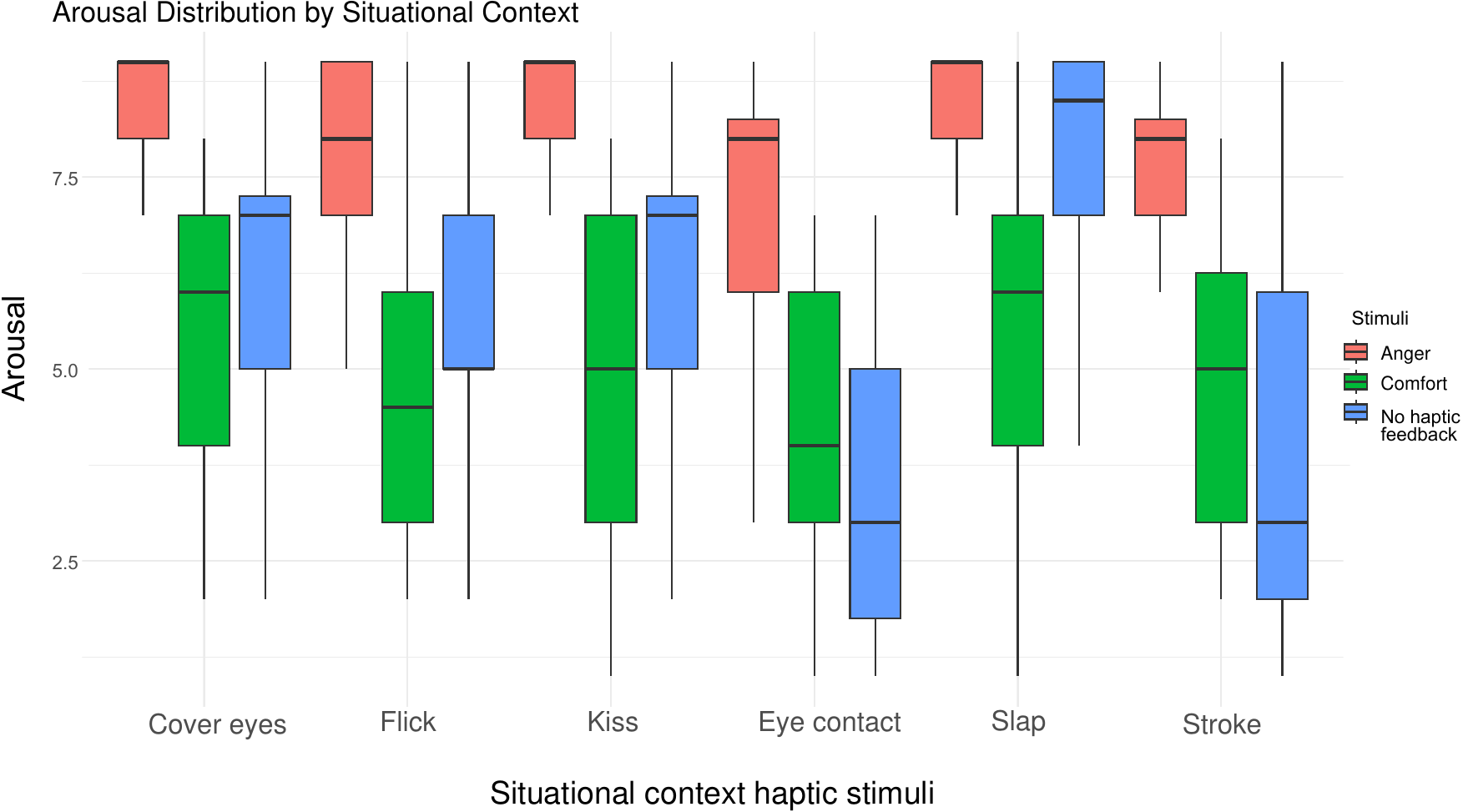}
\caption{Arousal comparison between the video only and video with tactile feedback, green bars represent ``Comfort'' stimulus, red bars are with ``Anger'' stimulus, and blue bars mean situational context video only, which is without haptic stimulus.}
    \label{fig:video_arousal}
\end{subfigure}
\begin{subfigure}{0.45\textwidth}
    \centering
    \includegraphics[width=\linewidth]{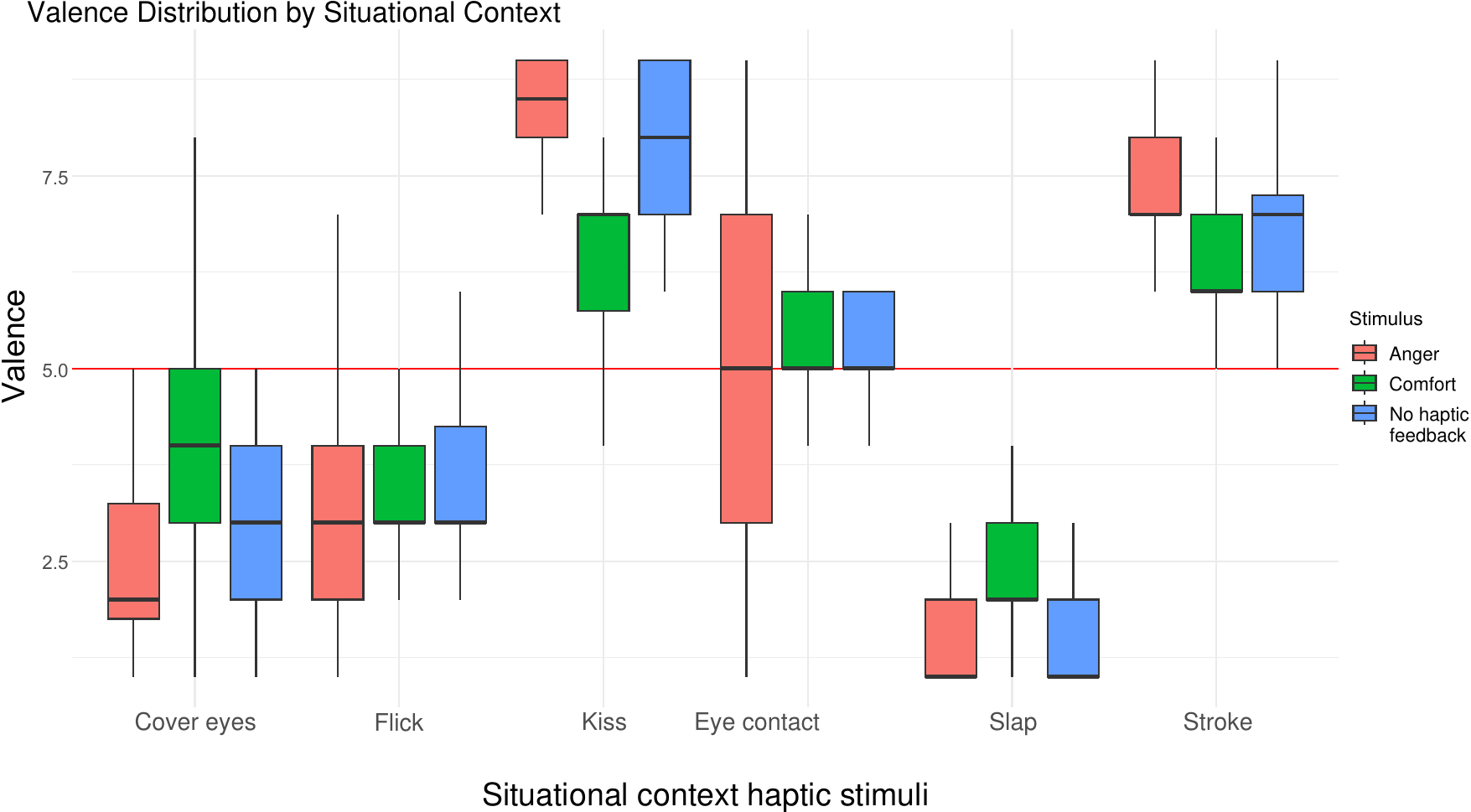}
    \caption{Valence comparison between the video only and video with tactile feedback, green bars represent ``Comfort'' stimulus, red bars are with ``Anger'' stimulus, and blue bars mean situational context video only, which is without haptic stimulus.}
    \label{fig:video_valence}
\end{subfigure}

\caption{Comparison between the video only and video with tactile feedback. Subfigue (a) shows that there is a significant difference between videos and videos with haptic stimulus on arousal. Subfigure (b) shows that the situational context valence overrides the haptic stimulus valence.}
\label{fig:video}
\end{figure*}


\subsection{Comparison between Haptic-Only and Video with Haptic Feedback}

\subsubsection{Arousal analysis}

We analyzed arousal levels in conditions involving action videos with haptic feedback and haptic-only scenarios (served as a no-situational context group). As shown in Fig.~\ref{fig:touch_arousal}, the Kruskal-Wallis test identified a significant main effect of situational context, which included six actions with haptic feedback and haptic-only conditions (where haptic-only conditions are considered as having no prior situational context), on arousal for both C\textsubscript{Tactile\textsuperscript{Anger}} ($H(6) = 20.18, p  = 0.003$) and C\textsubscript{Tactile\textsuperscript{Comfort}}($H(6) = 13.89, p = 0.03$). 

Post-hoc Wilcoxon signed-rank tests with Bonferroni correction revealed no significant differences in arousal levels between C\textsubscript{Tactile\textsuperscript{Comfort}} and six C\textsubscript{Context\textsuperscript{Comfort}}. Similarly, no significant differences were found between C\textsubscript{Tactile\textsuperscript{Anger}} and C\textsubscript{Context\textsuperscript{Anger}}. This finding suggests that tactile feedback can significantly override the arousal interpretation based on situational context. In addition, a significant difference was observed between the C\textsubscript{Slap\textsuperscript{Anger}} and the C\textsubscript{Stroke\textsuperscript{Anger}} ($p = 0.047$); C\textsubscript{Slap\textsuperscript{Anger}} and the C\textsubscript{Eye Contact\textsuperscript{Anger}} ($p = 0.041$). Moreover, only significant difference was found between C\textsubscript{Slap\textsuperscript{Comfort}} and the C\textsubscript{Eye Contact\textsuperscript{Comfort}} ($p = 0.041$). The number of significant differences observed among situational contexts with haptic feedback is notably lower than in the situational context-only condition, as illustrated in Fig.~\ref{fig:video_valence}. Additionally, the variation in arousal ratings across the six situational contexts is reduced when haptic feedback is introduced, converging toward the arousal level of the haptic stimulus. Furthermore, across all situational contexts with haptic feedback, a significant difference was found between actions featuring anger tactile feedback and those featuring comfort tactile feedback ($p<0.05$), suggesting that participants could distinguish between different arousal levels of haptic feedback in various action scenarios.

The results, summarized in Tab.~\ref{tab:arousal}, further indicate that tactile stimuli can override the arousal interpretation shaped by situational context, in addition, haptic feedback reduces variability in arousal ratings across different situational contexts, as shown in Fig.~\ref{fig:touch_arousal}.

\subsubsection{Valence analysis}

For valence, the results can be seen in Fig.~\ref{fig:touch_valence} and Table.~\ref{tab:valence}. The Kruskal-Wallis test identified a significant main effect of situational context on their valence level, which included six actions with haptic feedback and haptic-only conditions (where haptic-only conditions are considered as having no prior situational context), on valence for both C\textsubscript{Tactile\textsuperscript{Anger}} ($H(6) = 142.97, p < 0.001$) and C\textsubscript{Tactile\textsuperscript{Comfort}} ($H(6) = 106.90, p < 0.001$). Post-hoc Wilcoxon signed-rank tests with Bonferroni correction revealed significant differences in valence levels between C\textsubscript{Tactile\textsuperscript{Comfort}} and three action videos with C\textsubscript{Tactile\textsuperscript{Comfort}}, specifically C\textsubscript{Flick\textsuperscript{Comfort}} ($p=0.03$), C\textsubscript{Kiss\textsuperscript{Comfort}} ($p=0.03$), and C\textsubscript{Slap\textsuperscript{Comfort}} ($p<0.001$). Additionally, all six action pairwise comparisons showed significant differences ($p<0.05$), with the exception of C\textsubscript{Stroke\textsuperscript{Comfort}} and C\textsubscript{Kiss\textsuperscript{Comfort}}, where no significant difference was observed. Similarly, significant differences were found between C\textsubscript{Tactile\textsuperscript{Anger}} and the three C\textsubscript{Tactile\textsuperscript{Anger}}, including kiss ($p<0.001$), eye contact ($p<0.001$), and stroke ($p<0.001$). This indicates that tactile stimulus valence does not override the situational context valence.


\subsection{Comparison between Video-Only and Video with Haptic Feedback}

\subsubsection{Arousal analysis}


For arousal, we found a significant main effect of different haptic arousal levels (including conditions without haptic stimuli) across different actions ($p < 0.001$). Additionally, there was a significant difference in arousal ratings among all six C\textsubscript{Context\textsuperscript{Comfort}} ($p < 0.001$). For pairwise comparison, we found that arousal ratings were significantly higher in all C\textsubscript{Context\textsuperscript{Anger}} compared to C\textsubscript{Context} ($p < 0.001$). However, C\textsubscript{Context\textsuperscript{Comfort}}, significant differences were observed only in C\textsubscript{Stroke\textsuperscript{Comfort}} ($p = 0.03$) and C\textsubscript{Slap\textsuperscript{Comfort}} ($p < 0.001$) when compared to their respective video-only conditions.

\subsubsection{Valence analysis}

For valence, no significant differences were found between situational context with anger haptic stimuli and their corresponding situational context video-only conditions ($p > 0.05$). In addition, no significant differences were found between most situational context with comfort haptic stimuli and their corresponding video-only conditions, there are only two significant effects were observed, one for C\textsubscript{Kiss\textsuperscript{Comfort}} compared to C\textsubscript{Kiss} ($p < 0.001$), and another for C\textsubscript{Slap\textsuperscript{Comfort}} compared to C\textsubscript{Slap} ($p = 0.007$), which indicates that situational context overrides the valence interpretation derived from tactile feedback. 


As shown in Fig.~\ref{fig:touch_valence} and Fig.~\ref{fig:video_valence}, the situational context significantly overrides the valence interpretation derived from the tactile feedback. In addition, we observed that negative haptic stimuli (anger) tended to amplify emotions by reinforcing or mirroring the emotional intensity of the action, whereas positive haptic stimuli (comfort) appeared to modulate emotions, reducing the valence intensity of the experience, and tended to converge around the neutral valence rating. To examine how tactile feedback influences emotional responses, we compared situational context with haptic stimulus valence deviation from the neutral valence and used the original situational context valence as a reference. Then, we used a linear mixed-effects model. The model included \textit{Emotion Category} (C\textsubscript{Context} as reference level, C\textsubscript{Context\textsuperscript{Comfort}} and C\textsubscript{Context\textsuperscript{Anger}}) as a fixed effect and \textit{Participant ID (PN)} as a random intercept to account for inter-individual variability:

\vspace{-1.5em}

\begin{equation}
\text{Valence deviation} \sim \text{Valence Category} + (1 \mid \text{PN})
\end{equation}
\vspace{-1.5em}

Emotion deviation was defined as the absolute deviation from neutral valence (5), such that higher values reflect stronger emotional reactions regardless of direction.

Compared to the C\textsubscript{Context}, C\textsubscript{Context\textsuperscript{Comfort}} significantly reduced valence deviation ($\beta = -0.41$, $SE = 0.12$, $t(542) = -3.26$, $p = .001$), indicating that positive tactile stimulation moderated affective responses and brought them closer to neutrality. In contrast, C\textsubscript{Context\textsuperscript{Anger}} significantly increased valence deviation ($\beta = 0.58$, $SE = 0.12$, $t(542) = 4.68$, $p < .001$), suggesting that negative tactile stimulus amplified situational context valence.


The regulatory and amplifying effects of tactile stimulation on valence may be understood through Gross’s emotion regulation framework \cite{gross1998emerging}, which emphasizes the role of cognitive strategies in modulating emotional experiences. Positive touch likely engages cognitive reappraisal processes, enabling individuals to reinterpret emotional stimuli in a less extreme manner.  Conversely, negative touch may enhance emotional valence through increased emotional perception. This may involve emotion contagion—where tactile input intensifies affective resonance.


\section{Discussion}


This study explored the interaction between tactile feedback and situational context in shaping emotional interpretations during human-robot interaction. Firstly, we can confirm that participants could differentiate the arousal and valence from the solely haptic stimulus sent from the robot by a mediated vibration sleeve. 

For arousal, the findings demonstrate that tactile feedback can override situational context arousal ratings. In addition, Tactile stimulus decreased the variance among all the six situational contexts. One possible explanation is participants could feel the vibration amplitude from the haptic stimulus, which is difficult to be influenced by the situational context.

In addition, situational context plays a role in shaping valence interpretation under contexts with haptic stimulus, where contextual cues dominate the perception of emotion valence, leaving less room for haptic feedback to alter valence ratings. In addition, negative haptic stimuli (anger) amplify emotional responses, reinforcing the intensity of both positive and negative emotions, whereas positive haptic stimuli (comfort) moderate valence, shifting emotions toward a more neutral state, which could also be because of the high arousal state has interplay with the interpretation of the valence. These findings suggest that tactile feedback does not function in isolation but rather interacts with broader contextual cues to shape emotional responses.

In addition, visual context overwhelmingly dominated emotional interpretations in scenarios with strongly negative connotations, such as ``slap," leaving minimal space for tactile feedback to alter perceptions significantly. However, in scenarios with positive or ambiguous emotional connotations like ``kiss", ``stroke", or ``eye contact", tactile stimuli markedly influenced participants' emotional judgments, highlighting the importance of multimodal congruency in shaping emotional perceptions. In addition, participants decode ``eye contact'' with the robot as positive valence, which might caused by the appearance of the pepper robot or the eye interaction that could raise the positive emotion.

While the results underscore the significance of tactile-context congruency, some methodological limitations exist. The study was conducted in a controlled laboratory setting with video stimuli, potentially limiting ecological validity. Additionally, the homogeneous nature of the participant group restricts generalizability across diverse populations. Future research should examine similar multimodal emotional interactions in real-world settings and involve culturally diverse participant groups.

\section{Conclusion}

This study explored the interaction between tactile feedback and situational context in shaping emotional interpretations during human-robot interaction. The results indicate that situational context significantly dominates valence perception, with negative haptic stimuli (anger) amplifying overall valence level and positive haptic stimuli (comfort) moderating valence toward neutrality. Arousal was primarily driven by haptic feedback. These findings underscore the context-dependent nature of haptic feedback, reinforcing the need for adaptive, context-aware haptic systems in affective computing and human-robot interaction.

\bibliographystyle{ieeetr}
\bibliography{references}

\vfill

\end{document}